\documentclass[sigconf]{acmart} 
\renewcommand\footnotetextcopyrightpermission[1]{} 
\usepackage{adjustbox}

\usepackage{multirow} 
\usepackage{graphicx}
\usepackage[utf8]{inputenc} 
\usepackage[T1]{fontenc}    
\usepackage{hyperref}       
\usepackage{url}            
\usepackage{booktabs}       
\usepackage{amsfonts}       
\usepackage{nicefrac}       
\usepackage{microtype}      
\usepackage{xcolor}         
\usepackage{tabularx}
\usepackage{longtable}

\begin{document}

\title{See-Control: A Multimodal Agent Framework for Smartphone Interaction with a Robotic Arm}


\author{Haoyu Zhao}
\affiliation{%
  \institution{University College London}
  \city{London}
  \country{United Kingdom}}
\email{haoyu.zhao.25@ucl.ac.uk}

\author{Weizhong Ding}
\affiliation{%
  \institution{Imperial College London}
  \city{London}
  \country{United Kingdom}}
\email{dingweizhong1@gmail.com}

\author{Yuhao Yang}
\affiliation{%
  \institution{Huawei Noah’s Ark Lab}
  \city{London}
  \country{United Kingdom}}
\email{yang.yuhao@huawei.com}

\author{Zheng Tian}
\affiliation{%
  \institution{ShanghaiTech University}
  \city{Shanghai}
  \country{China}}
  \authornote{denotes corresponding author}
\email{tianzheng@shanghaitech.edu.cn}

\author{Linyi Yang}
\affiliation{%
  \institution{Southern University of Science and Technology}
  \city{Shenzhen}
  \country{China}}
\email{yangly6@sustech.edu.cn}

\author{Kun Shao}
\affiliation{%
  \institution{Huawei Noah’s Ark Lab}
  \city{London}
  \country{United Kingdom}}
\email{shaokun2@huawei.com}

\author{Jun Wang}
\affiliation{%
  \institution{University College London}
  \city{London}
  \country{United Kingdom}}
  \authornotemark[1]
\email{jun.wang@cs.ucl.ac.uk}

\begin{abstract}
Recent advances in Multimodal Large Language Models (MLLMs) have enabled their use as intelligent agents for smartphone operation. However, existing methods depend on the Android Debug Bridge (ADB) for data transmission and action execution, limiting their applicability to Android devices. In this work, we introduce the novel Embodied Smartphone Operation (ESO) task and present See-Control, a framework that enables smartphone operation via direct physical interaction with a low-DoF robotic arm, offering a platform-agnostic solution. See-Control comprises three key components: (1) an ESO benchmark with 155 tasks and corresponding evaluation metrics; (2) an MLLM-based embodied agent that generates robotic control commands without requiring ADB or system back-end access; and (3) a richly annotated dataset of operation episodes, offering valuable resources for future research. By bridging the gap between digital agents and the physical world, See-Control provides a concrete step toward enabling home robots to perform smartphone-dependent tasks in realistic environments.
\end{abstract}





\maketitle
\pagestyle{plain}
\section{Introduction}\label{sec:introduction}
State-of-the-art Multimodal Large Language Models (MLLMs), such as OpenAI o3 and GPT-4o \cite{hurst2024gpt}, have catalyzed a wave of task-oriented smartphone agents that automate mobile interfaces on behalf of users. However, most contemporary systems depend on Android Debug Bridge (ADB) for perception and control, tying solutions to a single OS, requiring developer modes and wired/debug privileges, and raising practical and privacy concerns outside lab settings. At the same time, embodied AI is rapidly moving from simulation to home environment \cite{yenamandra2023homerobot, latif2024physicsassistant}, where robots must complete daily tasks that routinely depend on mobile apps, such as approving two-factor prompts, scanning QR codes, navigating vendor-specific IoT controls, or coordinating deliveries. These converging trends open a timely Human-Computer Interaction (HCI) question: how can we enable reliable, privacy-preserving, \emph{cross-platform} smartphone operation that integrates naturally with the physical workflows of domestic robots? 

We argue that smartphone operation is not merely another Graphical User Interface (GUI) automation problem but a core capability for future home robots. Households already rely on smartphones as the control plane for services and devices; human users fluidly mix physical actions (moving the phone, aligning the camera, tapping) with digital steps (interpreting novel layouts, acknowledging pop-ups). A home robot that cannot act on a phone will frequently stall at the last mile of assistance. Giving robots a robust way to \emph{See and Control} on smartphones therefore expands the space of assistive scenarios, including accessibility support for users with motor impairments and privacy-sensitive tasks where on-device data should never be tunneled through debugging channels.

Prior smartphone agents make progress on reasoning \cite{zhang_appagent_2023,Agents2}, planning \cite{wang_mobile-agent-v2_2024,MMAC-Copilot,Mobileexperts}, and multimodal perception \cite{yang2025aria,wang2025mp,wangmobileagent}, but their reliance on ADB remains a brittle abstraction boundary: it is platform-locked (Android-only), privileges a development configuration (debug mode, tethering), and routes sensitive content through software bridges that are ill-suited to everyday domestic use. Emulators and OS toolkits sidestep some friction but fail to capture the physical constraints of real phones (screen glare, touch friction, alignment errors), which matter for embodied execution and user trust.
\begin{figure*}
    \centering
    \includegraphics[width=\linewidth]{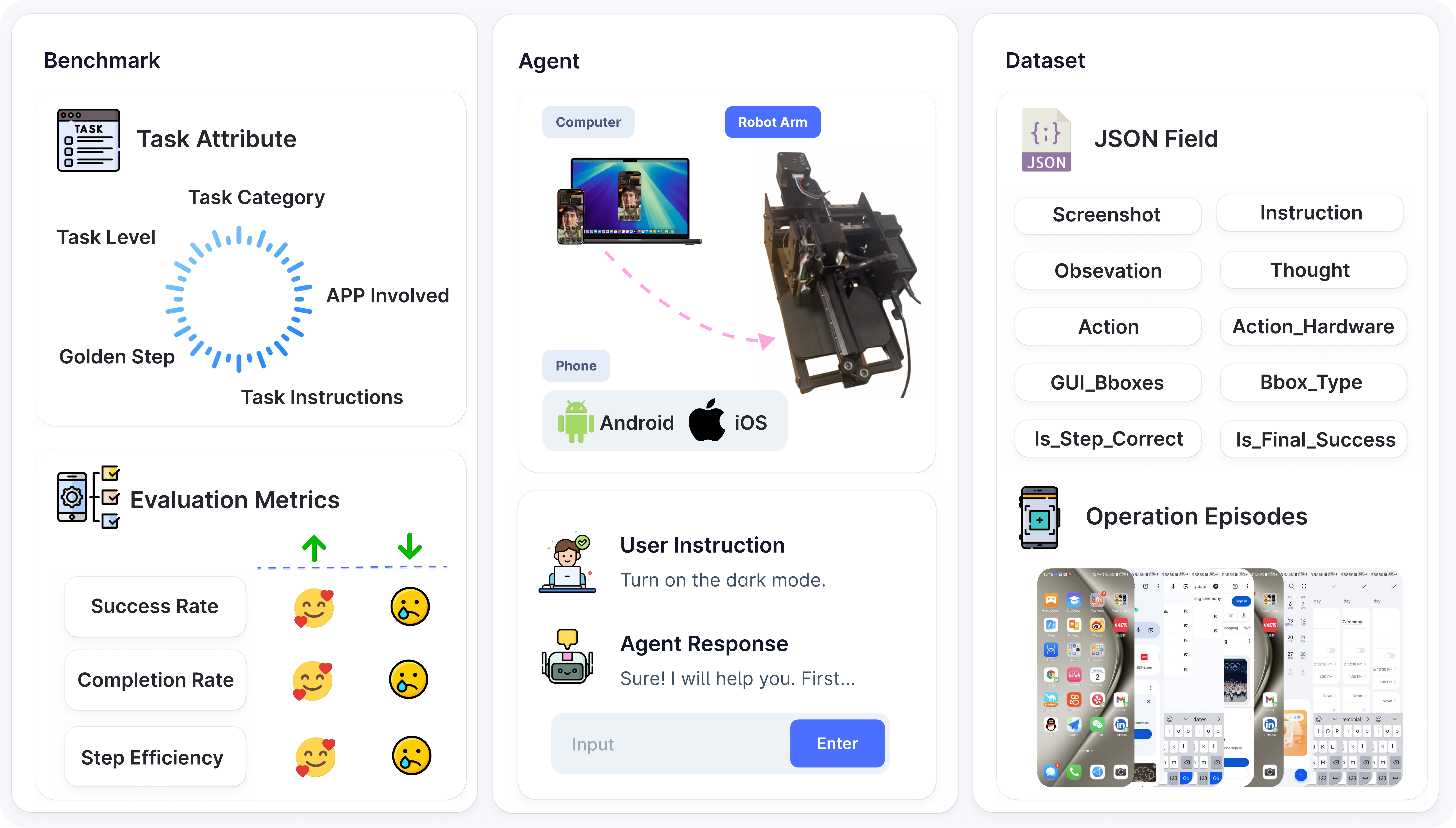}
    \caption{Overview of the See-Control Framework. The framework consists of three main components: a benchmark, an MLLM-based embodied agent, and a richly labeled dataset of smartphone operation episodes. See-Control reframes mobile HCI automation from a software-bridge problem into an embodied perception-and-action challenge, aligning with the way people naturally interact with phones. It not only overcomes the platform limitations present in existing research but also provides a more secure solution.}
     \Description{Overview of the See-Control framework for embodied smartphone operation. The left panel illustrates the benchmark, including task attributes (e.g., task category, task level, app involved, golden step, and task instructions) and evaluation metrics (success rate, completion rate, and step efficiency). The center panel shows the agent setup, where a computer interfaces with a low-DoF robot arm to physically interact with smartphones (Android/iOS). User instructions are issued in natural language, and the agent provides corresponding responses. The right panel presents the dataset structure, containing JSON fields (e.g., screenshot, instruction, observation, action, GUI bounding boxes, and correctness indicators) and operation episodes, which record step-by-step smartphone interaction sequences.}
    \label{fig:Overview}
\end{figure*}
In this paper, we introduce the Embodied Smartphone Operation (ESO) task and \emph{See-Control}, a framework that enables MLLM-based embodied agents to operate smartphones physically via a low-DoF robotic arm, which can tap, swipe, and type directly on the screen without ADB or privileged access. Our approach is platform-agnostic by design: it treats the phone as a visual-tactile object in the world, processes only screen imagery, and outputs continuous control to the end-effector. This reframes mobile HCI automation from a software-bridge problem into an embodied perception-and-action problem, aligning with how people themselves operate phones.

Concretely, See-Control comprises three components (see Fig. \ref{fig:Overview}). First, we establish an ESO benchmark of 155 manually annotated tasks with task-level metrics tailored to single-contact, three-axis robotic arms. Second, we build an MLLM-based embodied agent that takes only screen captures as input and generates control for a robotic arm, fully decoupled from ADB. Third, we release a richly annotated dataset of smartphone operation episodes, including intermediate agent states to facilitate future development of Vision–Language–Action (VLA) models.

Our contributions are as follows:
\begin{itemize}
\item A problem formulation and benchmark for \emph{Embodied Smartphone Operation} (ESO), targeting the realities of physical smartphone operation with low-DoF arms.
\item \emph{See-Control}, a platform-agnostic, privacy-preserving system that performs direct physical interaction with smartphones—no wired connections, device-specific code, or privileged modes.
\item An open, richly labeled dataset of operation episodes and an empirical evaluation across diverse apps demonstrating the feasibility and robustness of embodied, ADB-free smartphone control.
\end{itemize}

Together, these results suggest a complementary path for smartphone agents that better matches real-world domestic contexts: rather than contorting phones into developer configurations, let robots interact with them as people do—seeing the screen, deciding what to do, and touching it. This shift broadens deployment beyond a single OS, reduces privacy exposure, and integrates naturally with embodied assistance in the home.

\section{Related Work} \label{sec:related_work}
\subsection{Smartphone Operation Agents}
The goal of smartphone operation agents is to develop AI assistants capable of performing user-specified tasks within smartphone-oriented environments, with a strong focus on enhancing human-computer interaction to improve user experience and task efficiency. Early approaches \cite{VASTA,zhang2018robust,SUGILITE,cypher1993watch} focused mainly on programming by demonstration (PbD) systems, which allow users to automate tasks by simply performing them once or multiple times. However, these PbD systems are often constrained by their limited generalization ability and lack of autonomy.

Recently, state-of-the-art Multimodal Large Language Models (MLLMs), such as OpenAI o3, GPT-4o \cite{hurst2024gpt}, and the QwenVL series \cite{Qwen-VL, Qwen2-VL, Qwen2.5-VL}, have demonstrated promising progress in multi-turn dialogue and tool use. Leveraging these capabilities, agents built upon MLLMs have emerged as the dominant solution in the modern development of smartphone operation agents, where the MLLM serves as the central reasoning and decision-making component. By drawing on the vast knowledge acquired during pre-training, such agents can generate responses that are adaptive to a wide range of tasks and environments. 

A key factor of smartphone operation agents is the ability to accurately map language instructions to specific Graphical User Interface (GUI) elements. The methods for grounding these GUI elements can be broadly categorized into two approaches. The first category primarily focuses on GUI-related text analysis, leveraging textual files as auxiliary input to facilitate accurate understanding of the GUI. This approach involves utilizing existing GUI metadata (e.g., XML), with notable representative research like AppAgent \cite{zhang_appagent_2023} and AppAgent v2 \cite{li2024appagent}, or using GUI parsing modules to convert images into textual HTML files for further processing, as demonstrated in AutoDroid \cite{wen2024autodroid, wen2024autodroidv2}. The second category leverages computer vision tools, directly analyzing input images through a visual perception module, such as Grounding DINO \cite{groundingdino1} and CLIP \cite{li2022grounded}, to analyze screen information for determining the coordinates of interactive elements. This approach is exemplified by methods such as Mobile-Agent \cite{wangmobileagent}, and Mobile-Agent-E \cite{wang2025mobile}. Despite the variety of methods for GUI navigation, they all share a fundamental dependency on the Android Debug Bridge (ADB). This reliance not only introduces inherent security risks but also restricts agents' applicability to Android platforms, thereby limiting their cross-platform versatility.

\subsection{Smartphone Operation Benchmarks} 
The evaluation of agent performance in smartphone operation tasks is crucial for understanding their effectiveness, reliability, and adaptability in real-world scenarios. However, most existing agent frameworks are evaluated on a limited set of relatively simple tasks. To address this evaluation challenge, various benchmarks have been established, such as Spa-Bench \cite{Spa-bench}, Android World \cite{AndroidWorld}, Mobile Env \cite{Mobile-env}, Android Env \cite{AndroidEnv}, and Mobile-Bench \cite{Mobile-Bench}. Spa-Bench stands out as an outstanding example, offering a diverse set of tasks that cover both system and third-party apps in English and Chinese, thus bridging the gap in unified benchmarks by increasing task number and complexity. However, for the specific domain of Embodied Smartphone Operation (ESO), there remains a critical need for a benchmark that accounts for the unique characteristics of embodied interaction systems. Our work directly addresses this gap by introducing the first ESO benchmark.

\section{Benchmark} \label{sec:benchamrk}
In this section, we provide a comprehensive introduction to the See-Control benchmark. As the first evaluation framework specifically designed for the Embodied Smartphone Operation (ESO) task, the See-Control benchmark addresses the distinct challenges posed by using robotic arms with a single contact point to interact with smartphones.

\subsection{Task Construction}
In contrast to existing methods based on Android Debug Bridge (ADB), the ESO task presents a greater challenge for the agent, as the agent cannot invoke system commands via ADB and must accommodate cross-platform compatibility. Although most actions can be achieved through pixel coordinate mapping, three main actions present significant challenges that require additional adaptations, including navigating back, exiting applications, and typing. These actions can be completed easily with a single ADB command, but in the absence of ADB, they become substantially more complex. The See-Control benchmark is specifically designed to evaluate the performance of ESO agents in addressing these challenges, with tasks grounded in real user scenarios such as shopping, social interaction, and information management, ensuring that the benchmark is built upon common user needs. A list of the apps involved and their corresponding tasks can be found in Appendix \ref{Appendix}.

Specifically, we begin by compiling tasks from established smartphone operation frameworks and benchmarks, such as SPA-Bench \cite{Spa-bench}, and manually augment these tasks. In addition to augmenting existing tasks, we also manually design new tasks that specifically require the use of the three main challenging actions associated with the ESO task, thereby further enriching our evaluation task set. All tasks are then filtered to include only those that can be executed by robotic arms with a single contact point. Subsequently, we classified all tasks into two categories based on the number of applications required to complete each task: single-app tasks and cross-app tasks. Within the single-app tasks, these are further divided into two distinct groups. The first group includes tasks that only involve pixel-level coordinate mapping, which we refer to as the Standard Task Set. The second group consists of tasks that incorporate the aforementioned challenging actions, categorized under the Challenging Task Set. Given that cross-app tasks generally involve challenging actions, we do not differentiate among these tasks further. The benchmark comprises a total of 155 tasks, which are categorized into 140 single-app tasks and 15 cross-app tasks. Within the single-app tasks, 37 belong to the Standard Task Set, whereas the remaining 103 are part of the Challenging Task Set.

Furthermore, we categorize the tasks into 8 distinct categories according to their application domain, such as SystemApps, Comm \allowbreak\&\allowbreak Social, and others. Additionally, the benchmark emphasizes cross-platform generality, excluding tasks that are restricted to a single platform, such as Google Play, which is limited to the Android system. We further classify the single-app tasks into three levels of difficulty: easy, medium, and hard, with the majority of higher-difficulty tasks built upon their lower-level counterparts. For each task, we include a human-executed trajectory to identify the optimal execution path and determine the number of steps required, referred to as golden steps. For single-app golden step annotation, the calculation begins from the app's home screen. In contrast, for cross-app golden steps, which involve interactions across multiple applications, the calculation starts from the phone's home screen. Overall, our benchmark is designed to be cross-platform and fully customized for ESO tasks, enabling the systematic evaluation of home robots' ability to perform smartphone operations. These distinctive features set our benchmark apart from existing ones.

\subsection{Evaluation Metrics} 
To comprehensively evaluate the performance of ESO agents on our benchmark, we define three evaluation metrics that capture different aspects of task execution:(1) Success Rate (SR), (2) Completion Rate (CR), and (3) Step Efficiency (SE).
\begin{itemize}
    \item \textbf{Success Rate (SR): }A task is considered successfully executed when all components of the user’s instruction are fully completed. Let $S_n$ denote the total number of successfully executed tasks and $N$ the total number of tasks, then the success rate is defined as:
\begin{equation}
    SR = \frac{S_n}{N}
\end{equation}

    \item \textbf{Completion Rate (CR):} Although some instructions may not be fully executed, the successful actions performed by the agent still hold substantial importance. For a given task $T_i$, let $HStep_i$ represent the number of human steps required to complete the task, and $AStep_i$ the number of steps successfully completed by the agent. The Completion Rate for task $T_i$, denoted as $CR_{Ti}$, is then defined as:
\begin{equation}
    CR_{ti} = \frac{AStep_i}{HStep_i} \\
\end{equation}
    For a task set $S_i$ consisting of $n$ task, the overall Completion Rate $CR_{Si}$ is given by the average:
\begin{equation}
    CR_{Si} = \frac{1}{n} \sum_{i=1}^{n} CR_{ti}
\end{equation}
    
    \item \textbf{Step Efficiency (SE):} Assessing the efficiency of tasks successfully completed by the agent is another critical aspect. Evidently, a process that achieves the task goal with fewer steps reflects a more optimal solution. The Step Efficiency for task $T_i$, denoted as $SE_Ti$, is then defined as:
\begin{equation}
    SE_{ti} = \frac{HStep_i}{AStep_i} \\
\end{equation}
    For a task set $S_i$ consisting of $n$ task, the overall Step Efficiency $SE_{Si}$ is given by the average:
    \begin{equation}
    SE_{Si} = \frac{1}{n} \sum_{i=1}^{n} CR_{ti}
\end{equation}
\end{itemize}

\section{Agent}\label{sec:agent}
\begin{figure*}
    \centering
    \includegraphics[width=0.89\linewidth]{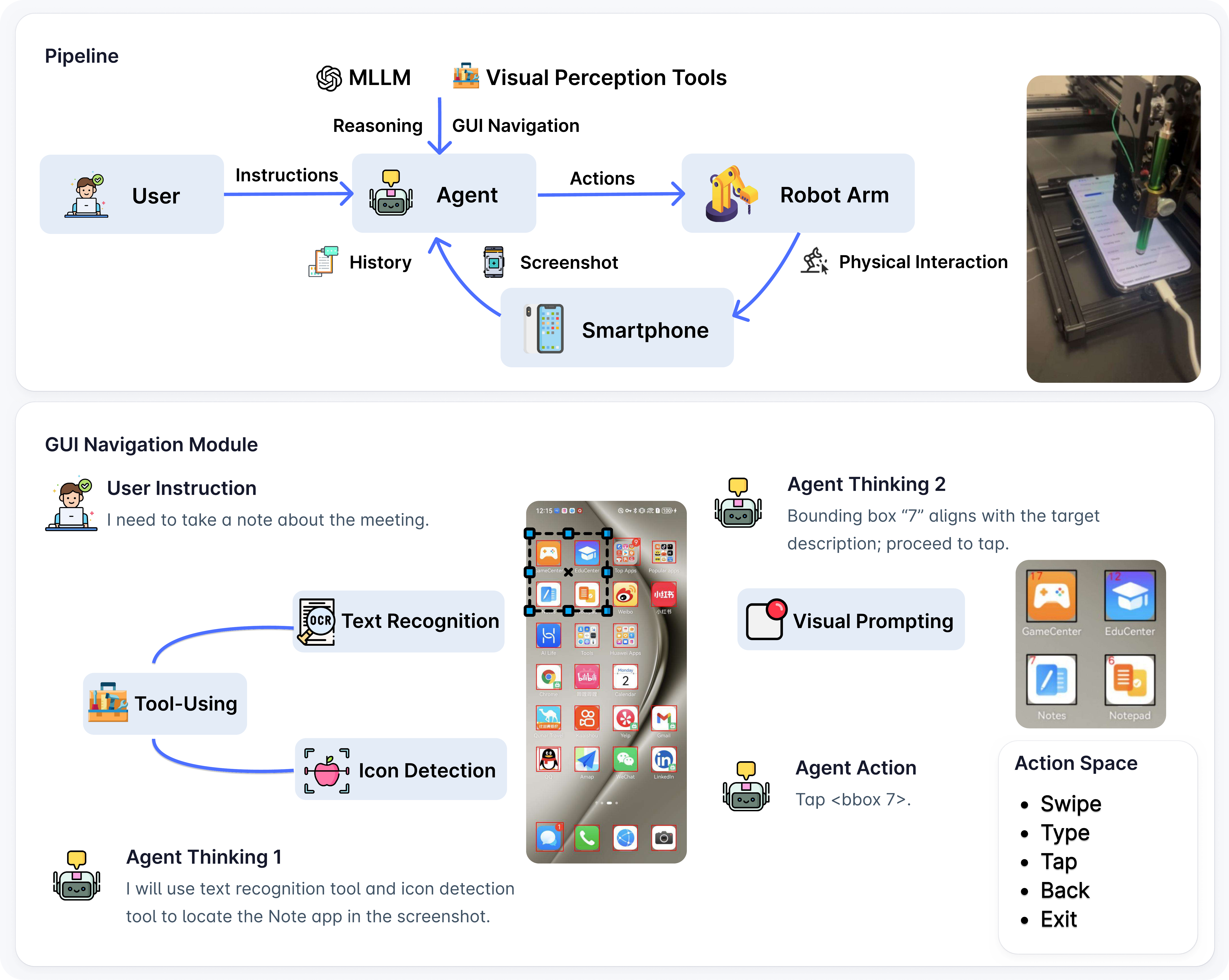}
    \caption{Pipeline of the See-Control Agent and its Visual Perception Module. The top panel shows how the agent processes user instructions and screen images to generate actions. The bottom panel provides a specific example of how the agent's reasoning leverages specialized visual grounding tools and visual prompting to accurately locate and interact with an UI element, demonstrating its ability to bridge user instructions with physical control.}
    \Description{System pipeline and GUI navigation module of the See-Control framework. The top panel illustrates the pipeline, where the user provides natural language instructions to the agent. The agent leverages Multimodal Large Language Models (MLLMs) and visual perception tools for reasoning and GUI navigation, sending action commands to a robot arm that physically interacts with the smartphone. The agent also receives screenshots and interaction history to refine its decisions. The right inset shows the robot arm executing a tap on a smartphone. The bottom panel details the GUI navigation module. Given a user instruction (e.g., taking a note about a meeting), the agent employs tool-using strategies such as text recognition and icon detection to locate the relevant app, supported by visual prompting. The agent generates intermediate reasoning steps (e.g., identifying bounding boxes) before executing an action (e.g., tapping a detected icon). The action space includes swipe, type, tap, back, and exit operations.}
    \label{fig:Pipeline}
\end{figure*}

\begin{figure*}
    \centering
    \includegraphics[width=0.8\linewidth]{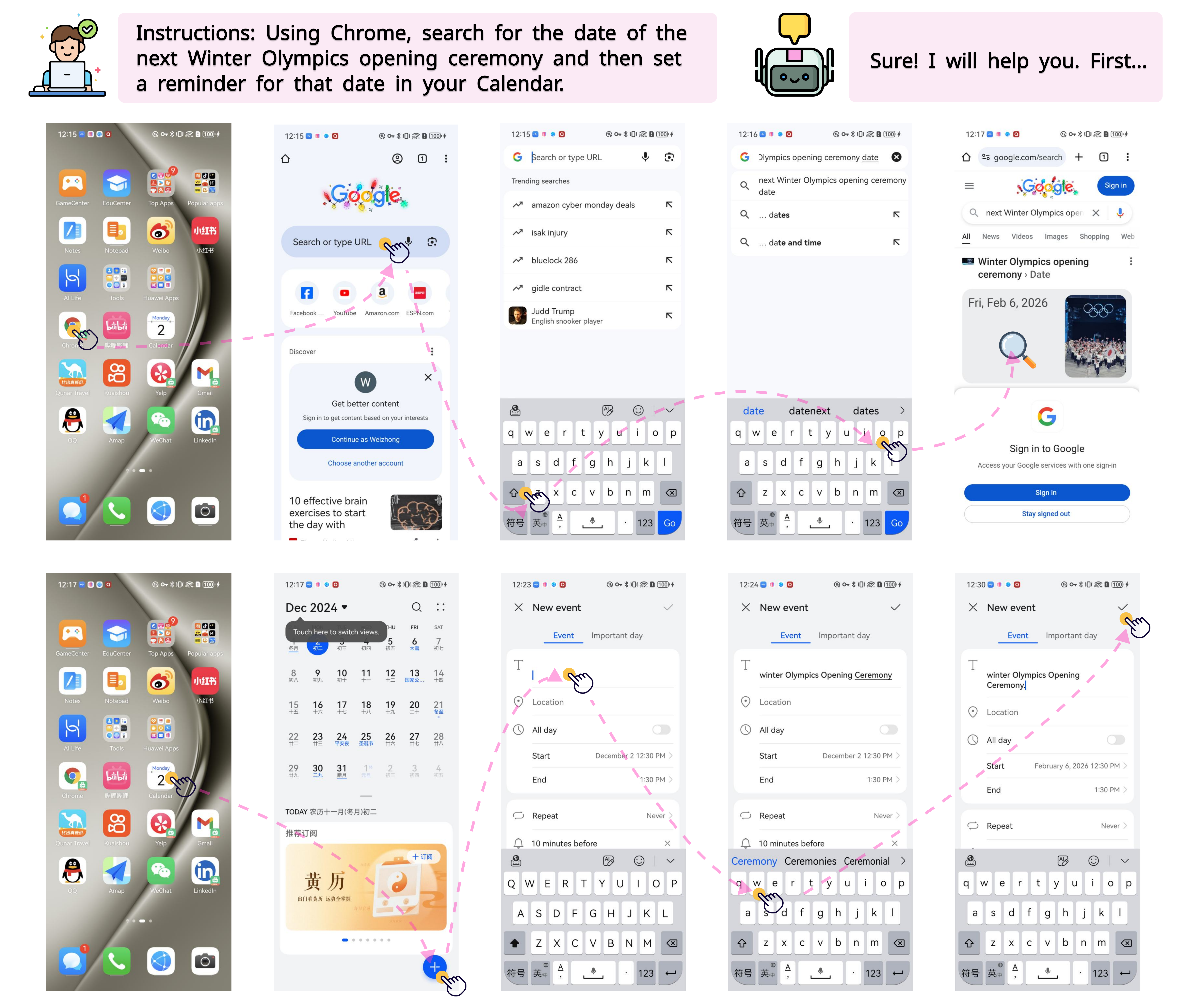}
    \caption{An example of the execution pathway of the See-Control agent during task completion, highlighting its comprehensive capabilities in task planning, reasoning, and visual perception.}
    \label{fig:route}
    \Description{Step-by-step illustration of a task execution pathway of the See-Control agent during task completion: (1) launching Chrome from the home screen; (2–5) searching in Google for the date of the next Winter Olympics opening ceremony; (6–7) opening the Calendar application; (8–11) creating a new event by entering the retrieved date (February 6, 2026) and setting a reminder. The dashed arrows indicate the sequential flow of agent actions.}
\end{figure*}

Designing an embodied smartphone operation agent presents a significant challenge due to several key factors. First, achieving cross-platform compatibility requires the system to avoid relying on system commands. For instance, actions like Text, Back, and Exit can be executed with a single command in ADB-based frameworks. However, without ADB, these operations require custom design and implementation strategies, making it challenging to achieve the same level of accuracy. The performance of these fundamental actions directly influences the overall accuracy of the embodied agent. Second, the robotic arm, with its single contact point, can only simulate single-finger operations. This hardware limitation necessitates the meticulous and specific design of every action's execution. Third, in the absence of ADB, the agent is inherently isolated from the smartphone’s internal storage, making it impossible to utilize internal GUI description files. Consequently, a fully vision-based approach is required to interpret and interact with the interface. In this work, we establish the simplest baseline for the Embodied Smartphone Operation (ESO) task, leaving the development of more efficient ESO agents as an open challenge for future research. An overview of our agent is shown in Fig. \ref{fig:Pipeline}, and Fig. \ref{fig:route} illustrates an example of the Execution Pathway of the See-Control agent during task completion.

\subsection{Framework}
\textbf{Hardware Setup.~}
Our hardware setup includes a PC, a three-axis robotic arm, and a smartphone, compatible with any operating system and screen resolution. The robotic arm is connected to the PC to receive hardware actions generated by the agent. The agent employs a fully vision-based approach, relying solely on real-time screenshots captured via screen mirroring. This system fully simulates the logic of human smartphone interaction, eliminating the need for cable connections to the smartphone or access to system files, such as XML, through software bridges.

\noindent\textbf{Action Space.~}\label{actionspace}
To reduce operational complexity, we define a fully discretized action space that closely mirrors typical human interactions with smartphones. This ensures that the set of actions is both minimal and sufficient to achieve the desired tasks. From the broad range of possible human actions, we select those that can be executed by home robots equipped with robotic arms featuring a single contact point. Each action and its corresponding function are represented in JSON format, with each action type corresponding to specific movements of the robotic arm. Below is a detailed description of the action space:

\begin{itemize}
    \item \textbf{Tap (\textit{x}, \textit{y}).}~~Perform a single tap at a specific pixel coordinate (\textit{x}, \textit{y}).
    \item \textbf{Swipe (\textit{x}, \textit{y}, \textit{direction}, \textit{distance})}. Swipe from the starting point at (\textit{x}, \textit{y}) in the given direction (\textit{up}, \textit{down}, \textit{left}, \textit{right}) over a specified distance (\textit{short}, \textit{medium}, \textit{long}). The distances are defined as 1/4, 1/3, and 1/2 of the screen's width or height, respectively.
    \item \textbf{Text (\textit{text})}.~~Input specific text into an on-screen text field.
    \item \textbf{Back}.~~Navigate to the previous screen.
    \item \textbf{Exit}.~~Close the current application or activity and return to the home screen.
\end{itemize}

\begin{figure*}
    \centering
    \includegraphics[width=\linewidth]{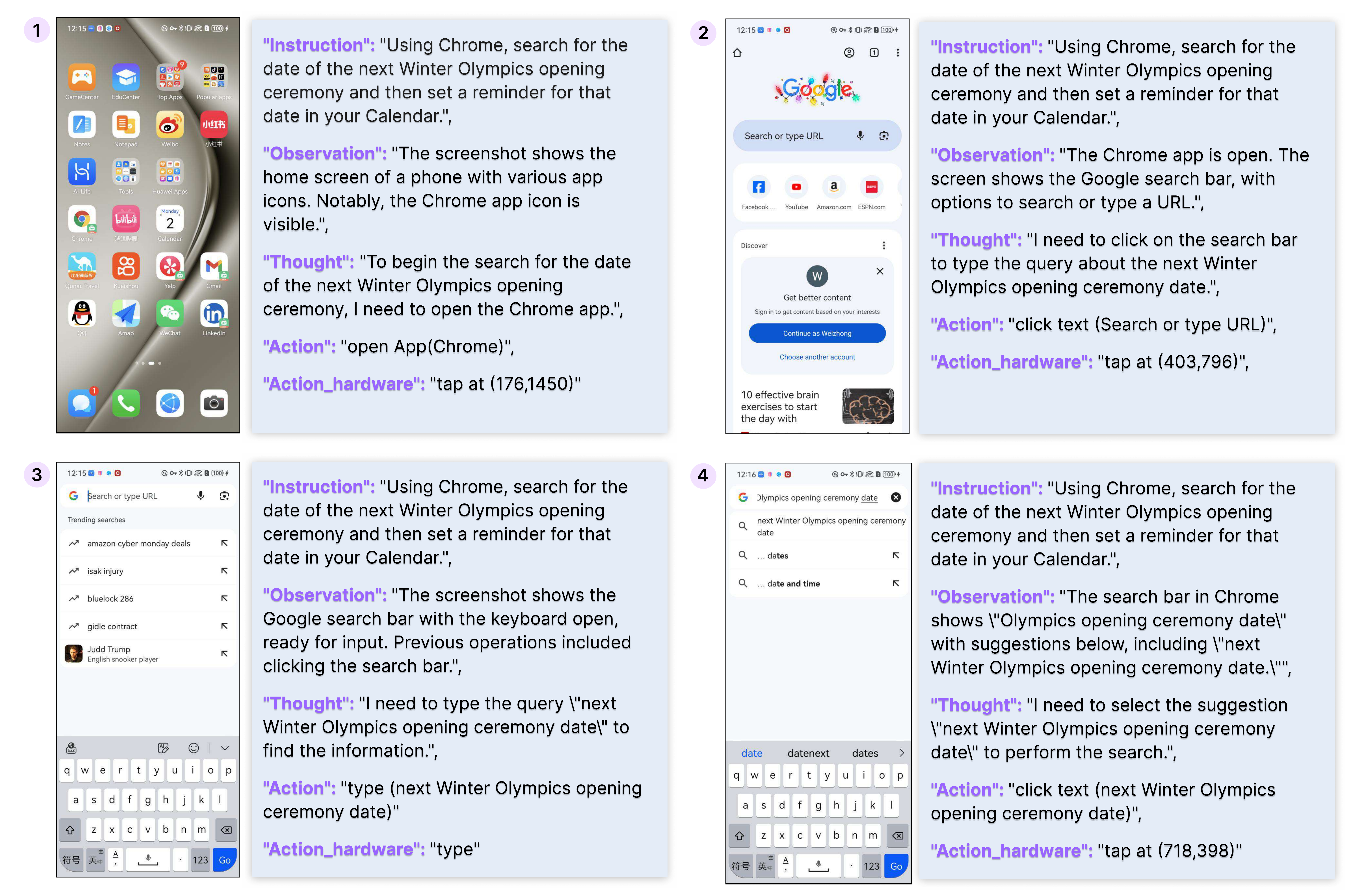}
    \caption{An example of the See-Control agent’s thinking and decision-making process, in which the agent interacts with the environment to gather observations for each state, performs reasoning, and converts its decisions into precise actions for the robotic arm. Due to space constraints, only the first four steps of the task are presented.}
    \label{fig:example}
    \Description{Example of an operation episode in the See-Control dataset. Each step includes a screenshot, paired with structured annotations: the user instruction, the agent’s observation of the interface, the agent’s intermediate thought process, the chosen action, and the corresponding hardware-level action. This example illustrates a multi-step interaction where the agent follows a natural language instruction—“search for the date of the next Winter Olympics opening ceremony using Chrome.” Steps 1–4 show how the agent sequentially (1) identifies and opens Chrome from the home screen, (2) selects the Google search bar, (3) types the query, and (4) chooses the suggested search completion, successfully retrieving the correct event date.}
\end{figure*}

\noindent\textbf{Agent Formulation.~}
Following Agent Q \cite{AgentQ}, we model the smartphone operation task as a Partially Observable Markov Decision Process (POMDP), given the limited observability of information beyond screenshots. Specifically, the agent lacks prior ground truth regarding interactive GUI elements on smartphones, thereby necessitating more precise localization of the interaction objectives. Furthermore, the smartphone screen represents a constantly evolving and highly dynamic environment, posing significant challenges for the agent in determining the exact state after the execution of an action.

We consider a generalized yet simplified POMDP setup, with an example of the thinking and decision-making process in the See-Control agent shown in Fig.~\ref{fig:example}. In this framework, the agent's observations, denoted as $\mathbf{o}_t \in \mathcal{O}$, encompass information provided by both the user and the smartphone. Specifically, the initial observation $\mathbf{o}_1$ includes the user's instruction and the current screenshot of the smartphone interface. The agent actions $\mathbf{a}_t \in \mathcal{A}$ are composite, based on the agent's memory component history, $\mathbf{h}_t$, where$\mathbf{h}_t = (\mathbf{a}_1, \ldots, \mathbf{a}_{t-1}, {\mathbf{o}_t })$. For the first action following the initial observation $\mathbf{o}_1$, we leverage the planning capabilities o of the MLLM to prompt the agent to generate a thought action $\mathbf{a}_t^{\text{tht}}\sim\pi(\mathbf{a}_t^{\text{tht}}|\mathbf{h}_t)$, which encapsulates the reasoning logic underlying the subsequent action. Subsequently, we generate the mobile interaction command $\mathbf{a}_t^{\text{env}}\sim\pi(\mathbf{a}_t^{\text{env}}|\mathbf{h}_t, \mathbf{a}_t^{\text{tht}})$, which corresponds to a tangible action such as "Tap", "Swipe", or "Text". We define the step action $\mathbf{a}_t$ as a tuple comprising observations, thought actions, and environment actions for all steps. During model optimization, we consider the joint likelihood to guide the decision-making process:
    \begin{align}  
    \label{matrix}
    \log \pi(\mathbf{a}_t|\mathbf{h}_t) =\log \pi(\mathbf{a}_t^{\text{env}}|\mathbf{h}_t, \mathbf{a}_t^{\text{tht}}) + \log \pi(\mathbf{a}^{\text{tht}}_t|\mathbf{h}_t)
    \end{align}

This formulation enables the agent to reason effectively about its actions while interacting with the highly dynamic smartphone environment. Building on this foundation, our agent is equipped with the capabilities of self-planning and self-reflection. Upon receiving a user task, the agent autonomously performs reasoning and planning to determine the subsequent action. Furthermore, it is capable of independently assessing whether the task has been completed and halting its output accordingly. In addition, the introduction of $\mathbf{h}_t$ enables the agent to perform self-reflection. Specifically, when the agent identifies that a particular action has resulted in an error, such as navigating to an incorrect page, it can revert to the previous step, modify the relevant action parameters, or generate alternative actions to autonomously correct its behavior.

\noindent\textbf{Gui Navigation.~}
In the See-Control agent, GUI elements are categorized into two distinct types: icons and text. Correspondingly, we have developed a dedicated visual perception module that allows agents to invoke grounding mechanisms tailored to each category. Additionally, we employ a visual prompting approach to enable the agent to perform more effective reasoning based on the visual marks provided. By analyzing the coordinates obtained from the GUI navigation process, we can accurately determine the actual execution trajectory of the robotic arm. 

For text grounding, consistent with most existing research, such as MobileAgent \cite{wangmobileagent} and AppAgent v2 \cite{AppAgentv2}, when the agent identifies the need to interact with text, it invokes an OCR model for detection, which returns the corresponding bounding boxes (bboxes). The center pixel of the bbox is then used as the contact point for the robotic arm to execute the desired action.

For icon grounding, inspired by LLM-Optic \cite{llmoptic} and SoM \cite{SoM}, we first employ a leading open-vocabulary object detector, Grounding DINO \cite{groundingdino1}, to detect all potential icon elements and return their corresponding bboxes. Subsequently, each candidate bbox is annotated with an index mark (See Fig. \ref{fig:Pipeline}), which serves as a visual prompt to enable the agent to perform more effective reasoning. The annotated image is then fed into the agent, which determines the specific icon index corresponding to the next action required for the task. This approach avoids over-reliance on basic computer vision models such as CLIP \cite{clip}, whose limited language understanding capabilities often lead to weak results in icon recognition. Instead, we leverage the outputs of the visual perception module in combination with the cross-modal semantic understanding capability of MLLMs to empower the agent to synthesize all available information and make direct decisions.

\noindent\textbf{\textbf{Action Execution.~}}
As mentioned in Sec. \ref{sec:benchamrk}, compared to existing agents that rely on system commands for smartphone operation, home robots
equipped with single-contact-point robotic arms face three main challenges due to their independence from the smartphone system: navigating back, exiting applications, and texting. In addition to these actions, which require special design, we also implement tap and swipe actions by mapping virtual coordinates to mechanical coordinates based on the pixel coordinates returned by the visual perception module. For the tap action, this involves clicking on the center of the bounding box, while for the swipe action, it entails dragging in a specified direction over a certain distance (as detailed in Sec. \ref{actionspace}).

\textit{Back and Exit.~~}For the back and exit actions, we primarily employ smartphone gestures as the main method, supplemented by GUI navigation-based approaches. In most scenarios, gestures can effectively serve as substitutes for the majority of system commands. For instance, swiping from the bottom to the top initiates the exit action, while swiping from left to right executes the back action. These gestures are particularly well-suited for a single-contact-point robotic arm, as they only require the implementation of a swipe motion from one coordinate to another. In the rare instances where gestures cannot successfully perform these actions, we adopt the same methodology utilized in GUI navigation. Specifically, we attempt to interact with the back button or relevant text elements to execute the desired actions.

\textit{Text.~~}In the ESO task, text input actions present non-trivial challenges. While ADB-based methods enable text input via system commands, achieving this without ADB requires additional design. In our agent, we address this challenge by implementing a keyboard key localization approach. By employing the standard international keyboard layout as an example, we systematically identify and store the pixel coordinates of each key. This localization procedure is performed only once for each device.

To achieve more precise localization, we first employ the icon grounding approach to identify the region where the keyboard is located. Following this, a cropping operation is applied to the corresponding section of the screenshot, isolating the relevant area for subsequent analysis. To reduce the computational complexity associated with localization, we do not localize every individual key. Instead, we utilize specific anchor points as references for positioning. Taking the first row of the keyboard as an example, we begin by identifying the positions of the first and second characters. By calculating the horizontal distance between these two anchor points, we can determine the positions of all characters in the first row. This is achieved by maintaining a constant vertical coordinate while incrementally adjusting the horizontal coordinate based on the calculated offset values. The same principle is then extended to determine the positions of keys in subsequent rows, ensuring accurate localization across the entire keyboard.

\subsection{Experiments}
We use GPT-4o \cite{hurst2024gpt} as the MLLM backbone for the See-Control agent and evaluate its performance using our carefully designed benchmark. The results are presented in Table \ref{tb:overall} to Table \ref{tb:crossapp}. All experiments are conducted in a unified environment, fully automated without human intervention. To ensure reliability, we save screenshots and actions at each step and perform secondary human verification to assess both the accuracy of each step and whether the task is successfully completed when the agent stops. This process ensures alignment with the intended human objectives.

\begin{table*}[!ht]
    \centering
    \caption{Overall Evaluation Results. SR = Success Rate, CR = Completion Rate, SE = Step Efficiency.}
    \resizebox{\textwidth}{!}{
    \begin{tabular}{l|ccc|ccc|ccc}
        \toprule
        \multirow{2}{*}{\textbf{Benchmark Subset}} & 
        \multicolumn{3}{c}{\textbf{Simple}} & 
        \multicolumn{3}{c}{\textbf{Medium}} & 
        \multicolumn{3}{c}{\textbf{Hard}} \\
        \cmidrule(lr){2-4} \cmidrule(lr){5-7} \cmidrule(lr){8-10}
        & SR $\uparrow$ & CR $\uparrow$ & SE $\uparrow$ & SR $\uparrow$ & CR $\uparrow$ & SE $\uparrow$ & SR $\uparrow$ & CR $\uparrow$ & SE $\uparrow$ \\
        \midrule
        Single-App Task Instructions (Standard Task Set) & 11/13 & 0.333 & 0.921 & 5/13 & 0.318 & 0.892 & 1/11 & 0.178 & 0.456 \\
        Single-App Task Instructions (Challenging Task Set) & 23/32 & 0.365 & 0.803 & 18/35 & 0.274 & 0.840 & 3/36 & 0.301 & 0.849 \\
        \midrule
        \multirow{2}{*}{Cross-App Task Instructions} 
        & \multicolumn{9}{c}{Overall Results} \\
        \cmidrule(lr){2-10}
        & \multicolumn{3}{c}{SR: 2/15} & \multicolumn{3}{c}{CR: 0.333} & \multicolumn{3}{c}{SE: 0.649} \\
        \bottomrule
    \end{tabular}
    }
    \label{tb:overall}
\end{table*}

From the results, we observe that the agent demonstrates proficiency in successfully completing most tasks within the simple instructions subset. However, as task complexity increases, the agent's Success Rate (SR) declines stepwise, with SR for cross-app tasks significantly lower than for single-app scenarios. This trend is also evident in the Completion Rate (CR) and Step Efficiency (SE). As task difficulty increases, the agent often requires more steps for trial and error, leading to a reduction in SE. At the same time, more complex tasks are often built upon simpler tasks, inherently requiring more steps. However, the agent often fails to recover from earlier mistakes, leading to a further drop in CR. 

The performance disparity between simple and complex tasks can be further attributed to two primary factors. First, our benchmark includes more complex tasks with a significantly higher average number of steps, which increases overall complexity and challenge. Longer task sequences require more GUI localization and interaction, inevitably reducing task execution accuracy. Second, unlike ADB-based methods, ESO settings present additional challenges. Simple tasks involve fewer ESO-specific actions, which reduces the likelihood of error accumulation.

\begin{table}[!ht]
    \centering
    \caption{Results for Single-App Task Instructions (Standard Task Set). " - " indicates Not Applicable, meaning that all tasks in the corresponding task category were either fully correct or fully incorrect. This convention also applies to the following table.}
    \resizebox{\linewidth}{!}{
    \begin{tabular}{l|ccc|ccc|ccc}
        \toprule
        \multirow{2}{*}{\textbf{Task Category}} & 
        \multicolumn{3}{c}{\textbf{Simple}} & 
        \multicolumn{3}{c}{\textbf{Medium}} & 
        \multicolumn{3}{c}{\textbf{Hard}} \\
        \cmidrule(lr){2-4} \cmidrule(lr){5-7} \cmidrule(lr){8-10}
        & SR & CR & SE & SR & CR & SE & SR & CR & SE \\
        \midrule
        Comm\&Social & 2/2 & - & 0.833 & 0/2 & 0.500 & - & 0/2 & 0.182 & - \\
        Lifestyle & 1/1 & - & 1.000 & 1/1  & - & 0.833 & 0/1 & 0.300 & - \\
       News\&Reading & 2/2 & - & 1.000 & 0/2 & 0.214 & - & 0/2 & 0.136 & - \\
       Prod\&Tools & 1/1 & - & 1.000 & - & - & - & - & - & - \\
        SystemApps & 4/4 & - & 0.867 & 4/5 & 0.167 & 0.906 & 1/3 & 0.256 & 0.455 \\
        Travel\&Nav & 1/3 & 0.333 & 1.000 & 0/3 & 0.317 & - & 0/3 & 0.111 & - \\
        \bottomrule
    \end{tabular}
    }
    \label{tb:standard}
\end{table}


\begin{table}[!ht]
    \centering
    \caption{Results for Single-App Task Instructions (Challenging Task Set).}
    \resizebox{\linewidth}{!}{
    \begin{tabular}{l|ccc|ccc|ccc}
        \toprule
        \multirow{2}{*}{\textbf{Task Category}} & 
        \multicolumn{3}{c}{\textbf{Simple}} & 
        \multicolumn{3}{c}{\textbf{Medium}} & 
        \multicolumn{3}{c}{\textbf{Hard}} \\
        \cmidrule(lr){2-4} \cmidrule(lr){5-7} \cmidrule(lr){8-10}
        & SR & CR & SE & SR & CR & SE & SR & CR & SE \\
        \midrule
        Comm\&Social & 7/10 & 0.267 & 0.716 & 3/10 & 0.378 & 0.953 & 2/10 & 0.232 & 0.774 \\
        Lifestyle & 3/3 & - & 0.722 & 4/4 & - & 0.721 & 1/5 & 0.304 & 1.000 \\
        Media\&Entmt & 4/5 & 0.667 & 0.838 & 3/5 & 0.238 & 1.000 & 0/5 & 0.248 & - \\
        News\&Reading & 3/4 & 0.667 & 0.833 & 1/4 & 0.185 & 1.000 & 0/4 & 0.199 & - \\
        Prod\&Tools & 2/2 & - & 0.667 & 1/3 & 0.293 & 0.700 & 0/3 & 0.371 & - \\
        Shop\&Fin & 2/3 & 0.667 & 0.700 & 2/3 & 0.143 & 0.586 & 0/3 & 0.524 & - \\
        SystemApps & 0/2 & 0.001 & - & 1/2 & 0.000 & 1.000 & 0/2 & 0.179 & - \\
        Travel\&Nav & 2/3 & 0.286 & 1.000 & 3/4 & 0.250 & 0.831 & 0/4 & 0.448 & - \\
        \bottomrule
    \end{tabular}
    }
    \label{tb:challenging}
\end{table}

\begin{table}[!ht]
    \centering
    \caption{Results for Cross-App Task Instructions.}
    \resizebox{\linewidth}{!}{
    \begin{tabular}{l|ccc}
        \toprule
        \textbf{Task Category} & 
        \textbf{SR} & 
        \textbf{CR} & 
        \textbf{SE} \\
        \midrule
        General Tool & 1/2 & 0.333 & 0.571 \\
        Information Management & 1/3 & 0.215 & 0.727 \\
        Media Entertainment & 0/1 & 0.545 & - \\
        Multi-Apps & 0/3 & 0.204 &  - \\
        Social Sharing & 0/3 & 0.252 & - \\
        Web Shopping & 0/3 & 0.553 & - \\
        \bottomrule
    \end{tabular}}
    \label{tb:crossapp}
\end{table}
Upon analyzing the execution trajectories, we identify the primary cause of task failures. Consistent with existing research, the main challenge stems from the inherent limitations of the visual perception module and the insufficient UI comprehension capabilities of MLLMs. Due to the diversity of UI designs, current models often struggle to interpret interfaces accurately. For instance, a "settings" button may appear in very different forms across applications, and in some cases, the correct action requires clicking on the user’s avatar. While such variations are trivial for users, they remain highly challenging for grounding models and MLLMs. These findings underscore the need for more robust visual perception mechanisms and the development of MLLMs with stronger autonomous reasoning abilities.

\subsection{User Study}
To further investigate users' acceptance, preferences, and concerns regarding smartphone operation agents, we conducted an exploratory user study with participants spanning a wide age range and diverse educational backgrounds. Each participant completed a questionnaire using a 5-point Likert scale, which focused on three key aspects: (1) attitudes toward existing smartphone operation agents, (2) concerns regarding privacy, security, and platform-specific limitations, and (3) expectations and preferences for future robotic smartphone automation, including perceived convenience. The survey was designed to be simple and intuitive, requiring no prior technical knowledge, and was administered anonymously. In total, 25 valid responses were collected. The findings indicate that the majority of users hold negative attitudes toward existing software-bridge-based agents, expressing reluctance to employ such methods in daily life (17 out of 25). Concerns were primarily centered on privacy and security risks (18 out of 25), as well as issues related to platform compatibility and usability (21 out of 25). Conversely, most users expressed a strong expectation that domestic intelligent robots should possess the capability to perform smartphone operation tasks (20 out of 25), with only a very small minority unwilling to delegate such tasks to home robots (1 out of 25). Furthermore, a large proportion of users believed that such an assistant would significantly enhance everyday convenience (22 out of 25). This user study further substantiates the value of See-Control in offering a complementary pathway for MLLM-based smartphone operation agents, particularly in broadening the design space for assistive human–computer interaction, enhancing accessibility support, and fostering trustworthy in-home automation.

\section{Dataset}\label{sec:dataset}
The See-Control dataset represents the first systematic effort to collect annotated data to support research on the Embodied Smartphone Operation task, capturing the episode history generated by the agent in a structured manner. As shown in Fig. \ref{fig:Overview}, the dataset comprises ten components, each of which is detailed in Table \ref{tab:attributes}. The variable \textit{IsStepCorrect} is determined by the state transition following each action, whereas \textit{IsFinalSuccess} is evaluated based on the overall success or failure of the task upon completion. Both of these variables undergo a rigorous double verification process by humans to ensure the high quality and reliability of the dataset.
\begin{table}[htbp]
\centering
\caption{Description of Attributes in the Dataset.}
\begin{tabularx}{\linewidth}{>{\raggedright\arraybackslash}p{3cm} X}
\toprule
\textbf{Attribute} & \textbf{Description} \\
\midrule
Screenshot & Screenshot of the smartphone display \\
Instruction & Task instruction provided by the user \\
Observation & Agent's observation of the current screen state \\
Thought & Agent's reasoning process for completing the task given the current screenshot and task instruction \\
Action & Abstract action decided by the agent, e.g., ``open App (Chrome)'' \\
Action Hardware & Actual hardware action executed by the robotic arm, e.g., ``tap at (176, 1450)'' \\
Gui Bboxes & Bounding boxes of all interactive elements detected by the visual perception module \\
Bbox Type & Categorical label for each bounding box, identifying the element type (e.g., icon or text) \\
IsStepCorrect & Whether the current step was performed correctly \\
IsFinalSuccess & Whether the overall task was completed successfully \\
\bottomrule
\end{tabularx}
\label{tab:attributes}
\end{table}

This dataset is expected to support three primary applications. First, it facilitates the training of MLLMs with enhanced GUI understanding. The inclusion of screenshots and their corresponding GUI bboxes enables the development of MLLMs capable of directly predicting the pixel coordinates of interactable elements, eliminating the need for additional visual perception modules. Second, the action sequences enable the training of Vision-Language-Action (VLA) models for direct robotic control, further reducing latency. Third, the dataset includes both successful and unsuccessful attempts, offering a diverse set of positive and negative examples. Each action step is annotated with success or failure labels, and the overall task outcome is documented. These detailed annotations provide a rich source of labeled data for training Process Reward Models (PRMs), which can enhance the model's reasoning capabilities through test-time scaling, further improving model performance.
\section{Limitations and Future Work}
While See-Control demonstrates the effectiveness of a novel embodied approach to smartphone interaction, it also faces several limitations. First, due to hardware constraints, the current design of See-Control supports only a simplified action space, limited to single-touch operations while excluding advanced controls such as multi-touch and complex gestures. This restriction may undermine the effectiveness of the agent in more complex interaction scenarios. Second, the latency inherent to MLLMs persists. In tasks with rapidly changing UIs, such as games, the combination of an MLLM with a visual perception module still suffers from model invocation delays, which may lead to execution errors due to dynamic environmental changes.

Addressing these limitations requires both hardware and algorithmic enhancements. Our current prototype intentionally uses a simple, low-DoF arm to establish a clear baseline and isolate core challenges in visual grounding, UI reasoning, and policy execution. A natural next step is to investigate dexterous end-effectors (multi-finger hands with compliant control and force sensing) and richer action spaces (precise multi-touch gestures, long-press, drag-and-drop), alongside tighter perception–action loops via end-to-end Vision Language Action (VLA) training and on-device inference for latency and privacy. We anticipate that advancing along these axes—dexterity, robustness, and efficiency—will enable embodied smartphone agents to interoperate with traditional app agents, letting systems fluidly choose the safest and most reliable pathway (software bridge vs. physical manipulation) for a given user, task, and environment.

\section{Conclusion}
In summary, Embodied Smartphone Operation (ESO) agents offer a complementary path, rather than a replacement, for MLLM-based smartphone operation agents. Software-based approaches, such as those using the Android Debug Bridge (ADB), remain effective for development and controlled settings, while our embodied route targets users and scenarios where safety, privacy, and cross-platform deployment are paramount, especially in domestic contexts where robots already act in the physical world. By formalizing the ESO problem and releasing \emph{See-Control}—a framework with a benchmark, baseline agent, and dataset—we provide the community with a reproducible scaffold to study perception-to-action pipelines that "see and touch" smartphones as people do, without privileged access and wired connections, and can be transferred to any touch-enabled device. This framing broadens the design space for assistive HCI, accessibility support, and trustworthy in-home automation.


\bibliographystyle{ACM-Reference-Format}
\bibliography{seecontrol}

@String{Computing = "Computing" }

@String{Computer = "{IEEE} Computer" }

@String{Chelsea = "Chelsea" }

@String{Springer = "Springer-Verlag" }

@inproceedings{li2022grounded,
  title={Grounded language-image pre-training},
  author={Li, Liunian Harold and Zhang, Pengchuan and Zhang, Haotian and Yang, Jianwei and Li, Chunyuan and Zhong, Yiwu and Wang, Lijuan and Yuan, Lu and Zhang, Lei and Hwang, Jenq-Neng and others},
  booktitle={Proceedings of the IEEE/CVF Conference on Computer Vision and Pattern Recognition},
  pages={10965--10975},
  year={2022}
}

@article{wang_mobile-agent-v2_2024,
  title={Mobile-agent-v2: Mobile device operation assistant with effective navigation via multi-agent collaboration},
  author={Wang, Junyang and Xu, Haiyang and Jia, Haitao and Zhang, Xi and Yan, Ming and Shen, Weizhou and Zhang, Ji and Huang, Fei and Sang, Jitao},
  journal={Advances in Neural Information Processing Systems},
  volume={37},
  pages={2686--2710},
  year={2024}
}

@article{hurst2024gpt,
  title={Gpt-4o system card},
  author={Hurst, Aaron and Lerer, Adam and Goucher, Adam P and Perelman, Adam and Ramesh, Aditya and Clark, Aidan and Ostrow, AJ and Welihinda, Akila and Hayes, Alan and Radford, Alec and others},
  journal={arXiv preprint arXiv:2410.21276},
  year={2024}
}

@misc{Qwen2.5-VL,
    title = {Qwen2.5-VL},
    url = {https://qwenlm.github.io/blog/qwen2.5-vl/},
    author = {Qwen Team},
    month = {January},
    year = {2025}
}

@article{Qwen2-VL,
  title={Qwen2-VL: Enhancing Vision-Language Model's Perception of the World at Any Resolution},
  author={Wang, Peng and Bai, Shuai and Tan, Sinan and Wang, Shijie and Fan, Zhihao and Bai, Jinze and Chen, Keqin and Liu, Xuejing and Wang, Jialin and Ge, Wenbin and Fan, Yang and Dang, Kai and Du, Mengfei and Ren, Xuancheng and Men, Rui and Liu, Dayiheng and Zhou, Chang and Zhou, Jingren and Lin, Junyang},
  journal={arXiv preprint arXiv:2409.12191},
  year={2024}
}

@article{Qwen-VL,
  title={Qwen-VL: A Versatile Vision-Language Model for Understanding, Localization, Text Reading, and Beyond},
  author={Bai, Jinze and Bai, Shuai and Yang, Shusheng and Wang, Shijie and Tan, Sinan and Wang, Peng and Lin, Junyang and Zhou, Chang and Zhou, Jingren},
  journal={arXiv preprint arXiv:2308.12966},
  year={2023}
}

@inproceedings{wangmobileagent,
  title={Mobile-Agent: Autonomous Multi-Modal Mobile Device Agent with Visual Perception},
  author={Wang, Junyang and Xu, Haiyang and Ye, Jiabo and Yan, Ming and Shen, Weizhou and Zhang, Ji and Huang, Fei and Sang, Jitao},
  booktitle={ICLR 2024 Workshop on Large Language Model (LLM) Agents}
}

@inproceedings{yang2025aria,
  title={Aria-UI: Visual Grounding for GUI Instructions},
  author={Yang, Yuhao and Wang, Yue and Li, Dongxu and Luo, Ziyang and Chen, Bei and Huang, Chao and Li, Junnan},
  booktitle={ICLR 2025 Workshop on Foundation Models in the Wild}
}

@article{wang2025mobile,
  title={Mobile-Agent-E: Self-Evolving Mobile Assistant for Complex Tasks},
  author={Wang, Zhenhailong and Xu, Haiyang and Wang, Junyang and Zhang, Xi and Yan, Ming and Zhang, Ji and Huang, Fei and Ji, Heng},
  journal={arXiv preprint arXiv:2501.11733},
  year={2025}
}

@inproceedings{zhang_appagent_2023,
  title={Appagent: Multimodal agents as smartphone users},
  author={Zhang, Chi and Yang, Zhao and Liu, Jiaxuan and Li, Yanda and Han, Yucheng and Chen, Xin and Huang, Zebiao and Fu, Bin and Yu, Gang},
  booktitle={Proceedings of the 2025 CHI Conference on Human Factors in Computing Systems},
  pages={1--20},
  year={2025}
}

@article{li2024appagent,
  title={Appagent v2: Advanced agent for flexible mobile interactions},
  author={Li, Yanda and Zhang, Chi and Yang, Wanqi and Fu, Bin and Cheng, Pei and Chen, Xin and Chen, Ling and Wei, Yunchao},
  journal={arXiv preprint arXiv:2408.11824},
  year={2024}
}

@inproceedings{wen2024autodroid,
  title={Autodroid: Llm-powered task automation in android},
  author={Wen, Hao and Li, Yuanchun and Liu, Guohong and Zhao, Shanhui and Yu, Tao and Li, Toby Jia-Jun and Jiang, Shiqi and Liu, Yunhao and Zhang, Yaqin and Liu, Yunxin},
  booktitle={Proceedings of the 30th Annual International Conference on Mobile Computing and Networking},
  pages={543--557},
  year={2024}
}

@article{wen2024autodroidv2,
  title={AutoDroid-V2: Boosting SLM-based GUI Agents via Code Generation},
  author={Wen, Hao and Tian, Shizuo and Pavlov, Borislav and Du, Wenjie and Li, Yixuan and Chang, Ge and Zhao, Shanhui and Liu, Jiacheng and Liu, Yunxin and Zhang, Ya-Qin and others},
  journal={arXiv preprint arXiv:2412.18116},
  year={2024}
}

@inproceedings{yenamandra2023homerobot,
  title={HomeRobot: Open-Vocabulary Mobile Manipulation},
  author={Yenamandra, Sriram and Ramachandran, Arun and Yadav, Karmesh and Wang, Austin S and Khanna, Mukul and Gervet, Theophile and Yang, Tsung-Yen and Jain, Vidhi and Clegg, Alexander and Turner, John M and others},
  booktitle={7th Annual Conference on Robot Learning}
}

@inproceedings{latif2024physicsassistant,
  title={Physicsassistant: An llm-powered interactive learning robot for physics lab investigations},
  author={Latif, Ehsan and Parasuraman, Ramviyas and Zhai, Xiaoming},
  booktitle={2024 33rd IEEE International Conference on Robot and Human Interactive Communication (ROMAN)},
  pages={864--871},
  year={2024},
  organization={IEEE}
}

@inproceedings{groundingdino1,
  title={Grounding dino: Marrying dino with grounded pre-training for open-set object detection},
  author={Liu, Shilong and Zeng, Zhaoyang and Ren, Tianhe and Li, Feng and Zhang, Hao and Yang, Jie and Jiang, Qing and Li, Chunyuan and Yang, Jianwei and Su, Hang and others},
  booktitle={European conference on computer vision},
  pages={38--55},
  year={2024},
  organization={Springer}
}

@inproceedings{Spa-bench,
  title={Spa-bench: A comprehensive benchmark for smartphone agent evaluation},
  author={Chen, Jingxuan and Yuen, Derek and Xie, Bin and Yang, Yuhao and Chen, Gongwei and Wu, Zhihao and Yixing, Li and Zhou, Xurui and Liu, Weiwen and Wang, Shuai and others},
  booktitle={NeurIPS 2024 Workshop on Open-World Agents},
  year={2024}
}

@inproceedings{AndroidWorld,
  title={AndroidWorld: A Dynamic Benchmarking Environment for Autonomous Agents},
  author={Rawles, Christopher and Clinckemaillie, Sarah and Chang, Yifan and Waltz, Jonathan and Lau, Gabrielle and Fair, Marybeth and Li, Alice and Bishop, William E and Li, Wei and Campbell-Ajala, Folawiyo and others},
  booktitle={The Thirteenth International Conference on Learning Representations}
}

@article{Mobile-env,
  title={Mobile-env: an evaluation platform and benchmark for LLM-GUI interaction},
  author={Zhang, Danyang and Xu, Hongshen and Zhao, Zihan and Chen, Lu and Cao, Ruisheng and Yu, Kai},
  journal={arXiv preprint arXiv:2305.08144},
  year={2023}
}

@misc{AndroidEnv,
      title={AndroidEnv: A Reinforcement Learning Platform for Android}, 
      author={Daniel Toyama and Philippe Hamel and Anita Gergely and Gheorghe Comanici and Amelia Glaese and Zafarali Ahmed and Tyler Jackson and Shibl Mourad and Doina Precup},
      year={2021},
      eprint={2105.13231},
      archivePrefix={arXiv},
      primaryClass={cs.LG},
      url={https://arxiv.org/abs/2105.13231}, 
}

@inproceedings{Mobile-Bench,
  title={Mobile-Bench: An Evaluation Benchmark for LLM-based Mobile Agents},
  author={Deng, Shihan and Xu, Weikai and Sun, Hongda and Liu, Wei and Tan, Tao and Liujianfeng, Liujianfeng and Li, Ang and Luan, Jian and Wang, Bin and Yan, Rui and others},
  booktitle={Proceedings of the 62nd Annual Meeting of the Association for Computational Linguistics (Volume 1: Long Papers)},
  pages={8813--8831},
  year={2024}
}

@misc{AgentQ,
      title={Agent Q: Advanced Reasoning and Learning for Autonomous AI Agents}, 
      author={Pranav Putta and Edmund Mills and Naman Garg and Sumeet Motwani and Chelsea Finn and Divyansh Garg and Rafael Rafailov},
      year={2024},
      eprint={2408.07199},
      archivePrefix={arXiv},
      primaryClass={cs.AI},
      url={https://arxiv.org/abs/2408.07199}, 
}

@misc{AppAgentv2,
      title={AppAgent v2: Advanced Agent for Flexible Mobile Interactions}, 
      author={Yanda Li and Chi Zhang and Wanqi Yang and Bin Fu and Pei Cheng and Xin Chen and Ling Chen and Yunchao Wei},
      year={2024},
      eprint={2408.11824},
      archivePrefix={arXiv},
      primaryClass={cs.HC},
      url={https://arxiv.org/abs/2408.11824}, 
}

@misc{llmoptic,
      title={LLM-Optic: Unveiling the Capabilities of Large Language Models for Universal Visual Grounding}, 
      author={Haoyu Zhao and Wenhang Ge and Ying-cong Chen},
      year={2024},
      eprint={2405.17104},
      archivePrefix={arXiv},
      primaryClass={cs.CV},
      url={https://arxiv.org/abs/2405.17104}, 
}

@misc{SoM,
      title={Set-of-Mark Prompting Unleashes Extraordinary Visual Grounding in GPT-4V}, 
      author={Jianwei Yang and Hao Zhang and Feng Li and Xueyan Zou and Chunyuan Li and Jianfeng Gao},
      year={2023},
      eprint={2310.11441},
      archivePrefix={arXiv},
      primaryClass={cs.CV},
      url={https://arxiv.org/abs/2310.11441}, 
}

@inproceedings{clip,
  title={Learning transferable visual models from natural language supervision},
  author={Radford, Alec and Kim, Jong Wook and Hallacy, Chris and Ramesh, Aditya and Goh, Gabriel and Agarwal, Sandhini and Sastry, Girish and Askell, Amanda and Mishkin, Pamela and Clark, Jack and others},
  booktitle={International conference on machine learning},
  pages={8748--8763},
  year={2021},
  organization={PmLR}
}

@inproceedings{VASTA,
  title={VASTA: a vision and language-assisted smartphone task automation system},
  author={Sereshkeh, Alborz Rezazadeh and Leung, Gary and Perumal, Krish and Phillips, Caleb and Zhang, Minfan and Fazly, Afsaneh and Mohomed, Iqbal},
  booktitle={Proceedings of the 25th international conference on intelligent user interfaces},
  pages={22--32},
  year={2020}
}

@inproceedings{SUGILITE,
  title={SUGILITE: creating multimodal smartphone automation by demonstration},
  author={Li, Toby Jia-Jun and Azaria, Amos and Myers, Brad A},
  booktitle={Proceedings of the 2017 CHI conference on human factors in computing systems},
  pages={6038--6049},
  year={2017}
}

@inproceedings{zhang2018robust,
  title={Robust annotation of mobile application interfaces in methods for accessibility repair and enhancement},
  author={Zhang, Xiaoyi and Ross, Anne Spencer and Fogarty, James},
  booktitle={Proceedings of the 31st Annual ACM Symposium on User Interface Software and Technology},
  pages={609--621},
  year={2018}
}

@book{cypher1993watch,
  title={Watch what I do: programming by demonstration},
  author={Cypher, Allen and Halbert, Daniel Conrad},
  year={1993},
  publisher={MIT press}
}

@article{MMAC-Copilot,
  title={MMAC-Copilot: Multi-modal Agent Collaboration Operating Copilot},
  author={Song, Zirui and Li, Yaohang and Fang, Meng and Li, Yanda and Chen, Zhenhao and Shi, Zecheng and Huang, Yuan and Chen, Xiuying and Chen, Ling},
  journal={arXiv preprint arXiv:2404.18074},
  year={2024}
}

@article{Mobileexperts,
  title={Mobileexperts: A dynamic tool-enabled agent team in mobile devices},
  author={Zhang, Jiayi and Zhao, Chuang and Zhao, Yihan and Yu, Zhaoyang and He, Ming and Fan, Jianping},
  journal={arXiv preprint arXiv:2407.03913},
  year={2024}
}

@article{Agents2,
  title={Agent s2: A compositional generalist-specialist framework for computer use agents},
  author={Agashe, Saaket and Wong, Kyle and Tu, Vincent and Yang, Jiachen and Li, Ang and Wang, Xin Eric},
  journal={arXiv preprint arXiv:2504.00906},
  year={2025}
}

@inproceedings{wang2025mp,
  title={Mp-gui: Modality perception with mllms for gui understanding},
  author={Wang, Ziwei and Chen, Weizhi and Yang, Leyang and Zhou, Sheng and Zhao, Shengchu and Zhan, Hanbei and Jin, Jiongchao and Li, Liangcheng and Shao, Zirui and Bu, Jiajun},
  booktitle={Proceedings of the Computer Vision and Pattern Recognition Conference},
  pages={29711--29721},
  year={2025}
}

\appendix
\section{Benchmark Details} \label{Appendix}
\subsection{APPs}
After conducting a thorough review of existing benchmarks and agent frameworks for smartphone operations, we identified a target set of 27 widely-used smartphone apps. The selection of these applications was based on their relevance to real-world usage scenarios, with a comprehensive description provided in Table~\ref{table:app-list}.
\begin{table*}[!ht]
\centering
\small
\renewcommand{\arraystretch}{1.5}
\caption{Categorized App List with Descriptions.}
\resizebox{\textwidth}{!}{%
\begin{tabular}{@{}l@{\hspace{10pt}}l@{\hspace{10pt}}l@{}}
\toprule
\textbf{Category} & \textbf{App name} & \textbf{Description} \\
\midrule
Comm\&Social & Facebook & A social networking platform to connect with friends, share updates, and join communities. \\
            & Gmail & Google's email service offering secure and efficient email management. \\
            & Instagram & A photo and video-sharing social networking platform. \\
            & LinkedIn & A professional networking platform for career development and business connections. \\
            & Reddit & A community-driven platform for discussions, news, and content sharing. \\
            & X (Twitter) & A microblogging platform for real-time updates, news, and social interactions. \\
\hline
Lifestyle & Deliveroo & A food delivery app offering meals from local restaurants. \\
        & Yelp & A platform for discovering and reviewing local businesses, including restaurants and services. \\
\hline
News\&Reading & BBC & A news app providing global and local news coverage from the BBC. \\
                & Quora & A question-and-answer platform where users can ask and answer questions on various topics. \\
\hline
Prod\&Tools & Chrome & A fast and secure web browser by Google. \\
            & Files & A file management app for organizing and accessing files on your device. \\
            & Microsoft OneNote & A note-taking app for capturing ideas, notes, and to-do lists. \\
            & Zoom & A video conferencing app for virtual meetings and webinars. \\
\hline
SystemApps & Calculator & A simple app for performing basic and advanced calculations. \\
            & Clock & An app for setting alarms, timers, and checking the time. \\
            & Settings & A system app for managing device settings and configurations. \\
            & Contacts & An app for managing and organizing your contact information. \\
\hline
Travel\&Nav & Airbnb & A platform for booking unique accommodations and experiences worldwide. \\
            & Booking & A travel app for booking hotels, flights, and rental cars. \\
            & Expedia & A comprehensive travel app for planning and booking trips. \\
\hline
Media\&Entmt & ESPN & A sports app for live scores, news, and video highlights. \\
            & Spotify & A music streaming app offering millions of songs and podcasts. \\
            & TikTok & A short-form video app for creating and sharing entertaining content. \\
            & YouTube & A video-sharing platform for watching and uploading videos. \\
\hline
Shop\&Fin & Amazon & An e-commerce app for shopping a wide range of products online. \\
            & Temu & A shopping app offering affordable products across various categories. \\
\bottomrule
\end{tabular}
} 
\label{table:app-list}
\end{table*}

\subsection{Tasks}
As demonstrated in Sec.~\ref{sec:benchamrk}, the benchmark consists of 155 tasks, categorized into 140 single-app tasks and 15 cross-app tasks. The detailed information for these tasks is provided in Table~\ref{table:standard_task_set} to Table~\ref{tab:cross_app_task_set}. 


\begin{table*}[h!]
\centering
\caption{Single-App Task Instructions (Standard Task Set) - 1.}
\label{table:standard_task_set}
\resizebox{\textwidth}{!}{ 
\begin{tabular}{ccccp{0.7\textwidth}}
\toprule
\textbf{Task Category} & \textbf{App Involved} & \textbf{Task Level} & \textbf{Golden Step} & \textbf{Task Instruction} \\
\midrule

\multirow{6}{*}{Comm\&Social}
& \multirow{3}{*}{Facebook} & 1 & 2 & Navigate to settings.\\
&  & 2 & 6 & Navigate to settings. Disallow notifications for Birthdays.\\
&  & 3 & 11 & Navigate to settings. Disallow notifications for Marketplace from Email and SMS. Disallow notifications for Memories from Email and SMS.\\
\cmidrule{2-5}
& \multirow{3}{*}{Gmail} & 1 & 2 & Navigate to settings.\\
&  & 2 & 6 & Navigate to settings. Check current setting for notifications. Turn off notification for Attachments.\\
&  & 3 & 11 & Navigate to settings. Check current setting for notifications. Turn off notification for Miscellaneous. Disable 'notification dot'. Return to Inbox.\\

\midrule
\multirow{3}{*}{Lifestyle}
& \multirow{3}{*}{Yelp} & 1 & 2 & Get the search results for nearby restaurants. \\
&  & 2 & 5& Get the search results for nearby restaurants. Filter to include only Chinese restaurants that offer takeout. Sort them by distance. \\
&  & 3 & 10 & Get the search results for nearby restaurants. Filter to include only Chinese restaurants that offer takeout. Sort them by distance. Select one result. Filter for 5-star reviews. \\

\midrule
\multirow{6}{*}{News\&Reading}
& \multirow{3}{*}{BBC} & 1 & 3 & Navigate to 'Innovation' section. Select 'Technology' tab. Open any news article. \\
&  & 2 & 10 & Go to app settings. Change the Text size to 'Smaller'. Navigate to 'Innovation' section. Select 'Technology' tab. Open any news article.\\
&  & 3 & 15 & Go to app settings. Change the Text size to 'Larger'. Navigate to 'Business' block. Select 'Technology of Business' tab. Open any news article. Save this article. Go to Saved Items to confirm the article was added. \\
\cmidrule{2-5}
& \multirow{3}{*}{Quora} & 1 & 3 & Discover any Space. Follow that space.\\
&  & 2 & 7 & Discover any Space. Follow that space. Go to questions in the space. Filter unanswered questions. Follow one question.  \\
&  & 3 & 11 & Discover any Technology Spaces. Follow that space. Also follow one of the suggested spaces. Turn off notification for the suggested space. Follow one of the contributors of the suggested space. \\

\midrule
\multirow{1}{*}{Prod\&Tools}
& Files & 1 & 3 & Go to the 'DCIM' folder in the internal storage.\\
\bottomrule

\end{tabular}
}

\end{table*}

\begin{table*}[h!]
\centering
\caption{Single-App Task Instructions (Standard Task Set) - 2.}
\label{table:standard_task_set1}
\resizebox{\textwidth}{!}{ 
\begin{tabular}{ccccp{0.7\textwidth}}
\toprule
\textbf{Task Category} & \textbf{App Involved} & \textbf{Task Level} & \textbf{Golden Step} & \textbf{Task Instruction} \\
\midrule
\multirow{18}{*}{SystemApps}
& \multirow{3}{*}{Calculator} & 1 & 4 & Get the result for '1+1'. \\
&  & 2 & 10 & Get the result for 'log(20)+ln(e)'.\\
&  & 3 & 14 & Get the result for 'log(20)+ln(e)'. Clear the results. Get the result for factorial 7.\\
\cmidrule{2-5}
& \multirow{6}{*}{Clock} & 1 & 4 & Set an alarm for 8am.\\
&  & 2 & 8& Set an alarm for 9am on weekdays. \\
&  & 3 & 15 & Set an alarm for 10am on weekdays. Disable vibration for this alarm. Set another alarm for 11am on weekends.\\
&  & 1 & 2& Add current time at London (UK) to clock. \\
&  & 2 & 6 & Set Home time zone to 'Hong Kong'. \\
&  & 3 & 12 & Add current time at Melbourne (Australia) to clock. Change style to Analog for clock. Change style to Analog for screen saver. \\
\cmidrule{2-5}
& \multirow{4}{*}{Settings} & 1 & 1& Check the current screen timeout.  \\
&  & 2 & 3& Check the current screen timeout. Set it to 5 minutes.  \\
&  & 3 & 7& Check the current screen timeout. Set it to 10 minutes. Then turn the dark theme on.   \\

\midrule
\multirow{9}{*}{Travel\&Nav}
& \multirow{3}{*}{Booking} & 1 & 3 & Navigate to app settings.  \\
&  & 2 & 7 & Navigate to app settings. Change Temperature to 'Degrees in Celsius'. Change Units to 'Metric (km, m)'.  \\
&  & 3 & 12 & Navigate to app settings. Change Currency to 'Pound Sterling'. Disable all notifications.  \\
\cmidrule{2-5}
& \multirow{6}{*}{Google Maps} & 1 & 3 & Get the search results for nearby hotel rooms.\\
&  & 2 & 7 & Get the search results for nearby hotel rooms. Filter the results to show only those that can accommodate 4 adults.  \\
&  & 3 & 10 & Get the search results for nearby hotel rooms. Filter the results to show only those that can accommodate 4 adults. Further filter the results with ratings higher than 4. \\
&  & 1 & 3 & Get the search results for nearby gas stations.  \\
&  & 2 & 6 & Get the search results for a nearby gas station that is open now. Get a driving route to it. \\
&  & 3 & 12 & Get the search results for a nearby gas station that is open now. Get a driving route with the gas station as the first stop. Set McDonald's as the final destination. \\
\bottomrule

\end{tabular}
}

\end{table*}

\begin{table*}[h!]
\centering
\caption{Single-App Task Instructions (Challenging Task Set) - 1.}
\label{table:challenging_task_set}
\resizebox{\textwidth}{!}{ 
\begin{tabular}{ccccp{0.7\textwidth}}
\toprule
\textbf{Task Category} & \textbf{App Involved} & \textbf{Task Level} & \textbf{Golden Step} & \textbf{Task Instruction} \\
\midrule

\multirow{32}{*}{Comm\&Social}
& \multirow{3}{*}{Facebook} & 1 & 3& Create a new post saying 'Hello World!'. \\
&  & 2 & 5& Create a new Public post saying 'Morning!'. Change to black background. \\
&  & 3 & 10& Create a new Public post saying 'Bonne Nuit'. Add the location as Eiffel Tower.  \\
\cmidrule{2-5}
& \multirow{3}{*}{Gmail} & 1 & 5 & Draft an email to agent.benchmark.2024@gmail.com asking them about their new paper.  \\
&  & 2 & 9 & Send an email to agent.benchmark.2024@gmail.com asking them about their new paper. Navigate to the Sent tab. Check the email details for confirmation after sending.  \\
&  & 3 & 11 & Draft an email to agent.benchmark.2024@gmail.com asking them about their new paper. Schedule it to be sent tomorrow morning. Navigate to the Scheduled tab. Check the email details for confirmation for confirmation after scheduling.  \\
\cmidrule{2-5}
& \multirow{6}{*}{Instagram} & 1 & 4 & Get the search results for 'Messi'.  \\
&  & 2 & 6 & Get the search results for 'Cristiano Ronaldo'. Follow one account. \\
&  & 3 & 10 & Get the search results for 'Minions'. Follow one account. Set to get all notifications when they goes live. Turn on notifications for their posts. \\
&  & 1 & 3 & Navigate to the page to edit my profile. \\
&  & 2 & 10 & Navigate to the page to edit my profile. Add bio 'Hello World!'. Change pronouns to 'it'. \\
&  & 3 & 17 & Navigate to the page to edit my profile. Add link 'https://github.com'. Change gender to Custom 'Them'. Switch to private account. \\
\cmidrule{2-5}
& \multirow{6}{*}{LinkedIn} & 1 & 4 & Get the search results for 'OpenAI'. Follow their page.  \\
&  & 2 & 7 & Get the search results for 'Huawei'. Follow their page. Filter the search results to Groups. Join one of the Huawei groups. \\
&  & 3 & 12 & Get the search results for 'Huawei HKRC'. Follow their page. Leave a 'Cheers!' comment on one of its posts. Like the post. Repost the post instantly. View the repost to confirm. \\
&  & 1 & 4 & Get the search results for 'Engineer' job.  \\
&  & 2 & 7 & Get the search results for 'Engineer' job in Spain. \\
&  & 3 & 10 & Get the search results for 'Engineer' jobs in Spain. Save one of the jobs. Confirm it is saved in My Jobs.  \\
\cmidrule{2-5}
& \multirow{3}{*}{Reddit} & 1 & 4 & Get the search results for 'r/worldnews'. Join the group.\\
&  & 2 & 8 & Get the search results for 'r/PremierLeague'. Filter posts for Liverpool. Join the group. Click on one of the posts. \\
&  & 3 & 11 & Get the search results for 'r/BlackMythWukong'. Join the group. Set community alerts to frequent. Click on one of the posts.  \\
\cmidrule{2-5}
& \multirow{6}{*}{X} & 1 & 3 & Draft a post with the content 'Written by Agent1'. \\
&  & 2 & 9 & Create a post with the content 'Written by Agent2'. Tag '\#animalcrossing'. Post it. Check it from the profile. \\
&  & 3 & 12 & Create a post with the content 'Written by Agent3'. Tag '\#animalcrossing'. Post it. Check it from the profile. Then Like it. Reply to it with 'Amazing post'. \\
&  & 1 & 5 & Search for the account @Mayday EN. Follow it.  \\
&  & 2 & 8 & Search for the account @Nintendo. Follow it. Search its post about 'Super Mario'. \\
&  & 3 & 15 & Search for the account @animalcrossing. Follow it. Search its post about 'Timmy and Tommy'. Repost one result. Check it from the profile for confirmation. \\
\bottomrule

\end{tabular}
}

\end{table*}


\begin{table*}[h!]
\centering
\caption{Single-App Task Instructions (Challenging Task Set) - 2.}
\label{table:challenging_task_set}
\resizebox{\textwidth}{!}{ 
\begin{tabular}{ccccp{0.7\textwidth}}
\toprule
\textbf{Task Category} & \textbf{App Involved} & \textbf{Task Level} & \textbf{Golden Step} & \textbf{Task Instruction} \\
\midrule
\multirow{12}{*}{Lifestyle}
& \multirow{9}{*}{Deliveroo} & 1 & 2 & Get the search results for McDonald's.  \\
&  & 2 & 5 & Get the search results for McDonald's. Enter a McDonald's restaurant. Search for fries there.  \\
&  & 3 & 10 & Get the search results for McDonald's. Enter a McDonald's restaurant. Search for fries there. Add a small fries to the basket. Add two medium fries to the basket. View the basket for confirmation. \\
&  & 2 & 5 & Get the search results for McDonald's. Enter two different McDonald's restaurant. \\
&  & 3 & 7 & Get the search results for McDonald's. Enter a McDonald's restaurant. Go back and search for KFC. Enter a KFC restaurant.  \\
&  & 3 & 12 & Get the search results for McDonald's. Enter a McDonald's restaurant. Search for fries there. Add a small fries to the basket. Go back and search for KFC. Enter a KFC restaurant. Search for fries there. Add a small fries to the basket.  \\
&  & 1 & 3 & Check Account details.  \\
&  & 2 & 5 & Check Account details. Return to home page. Get the search results for Starbucks.\\
&  & 3 & 12 & Get the search results for Starbucks. Enter a Starbucks restaurant. Add Starbucks to Favourites. Return to home page. Check Favourites.  \\
\cmidrule{2-5}
& \multirow{3}{*}{Yelp} & 1 & 2 & Get the search results for Burgers.  \\
&  & 2 & 5 & Get the search results for Chips. Get the search results for Burgers. \\
&  & 3 & 12 & Get the search results for Chips. Check one of the restaurants. Save to collection. Get the search results for Burgers.  \\

\midrule
\multirow{15}{*}{Media\&Entmt}
& \multirow{3}{*}{ESPN} & 1 & 3 & Get the search results for 'Klay Thompson'.  \\
&  & 2 & 5 & Get the search results for 'Klay Thompson'. See all the articles. Open one of the articles.\\
&  & 3 & 11 & Get the search results for 'Klay Thompson'. See all the articles. Open one of the articles. Return to the player's search results. Select the player. Turn on player news notification. Follow the player.  \\
\cmidrule{2-5}
& \multirow{3}{*}{Spotify} & 1 & 3 & Get the search results for the artist Taylor Swift.  \\
&  & 2 & 6 & Get the search results for the artist Taylor Swift. Enter her artist page. Shuffle play her playlist. \\
&  & 3 & 15 & Get the search results for the song 'Love Story' by Taylor Swift. Add this song to the new playlist namely 'Agent Playlist'. Then add another song 'The Scientist' by Coldplay to the same playlist. Check the playlist for confirmation.  \\
\cmidrule{2-5}
& \multirow{3}{*}{TikTok} & 1 & 3 & Get the search results for videos about pet cats.  \\
&  & 2 & 8 & Get the search results for videos about pet cats. Comment on a video with 'Such a cute cat.'  \\
&  & 3 & 13 & Get the search results for videos about pet cats. Comment on a video with 'Such a cute cat.' Swipe through another two videos and like them.  \\
\cmidrule{2-5}
& \multirow{6}{*}{YouTube} & 1 & 4 & Get the search results for the channel '@Tesla'. Subscribe to the channel.  \\
&  & 2 & 8 & Get the search results for the channel '@BMW'. Subscribe to the channel. Get the search results for the channel '@Mercedes'. Subscribe to the channel.  \\
&  & 3 & 12 & Get the search results for the channel '@Google'. Subscribe to the channel. Get the search results for the channel '@Microsoft'. Subscribe to the channel. Navigate to the Subscriptions tab. Show all subscriptions. Sort the subscriptions from A to Z. \\
&  & 1 & 3 & Get the search results for videos about LeBron James.  \\
&  & 2 & 10 & Get the search results for videos about LeBron James. Filter videos under 4 minutes.  \\
&  & 3 & 14 & Get the search results for videos about LeBron James. Filter videos under 4 minutes. Select any one of the results. Leave a comment 'great performance!'.  \\
\bottomrule

\end{tabular}
}

\end{table*}


\begin{table*}[h!]
\centering
\caption{Single-App Task Instructions (Challenging Task Set) - 3.}
\label{table:challenging_task_set}
\resizebox{\textwidth}{!}{ 
\begin{tabular}{ccccp{0.7\textwidth}}
\toprule
\textbf{Task Category} & \textbf{App Involved} & \textbf{Task Level} & \textbf{Golden Step} & \textbf{Task Instruction} \\
\midrule
\multirow{12}{*}{News\&Reading}
& \multirow{9}{*}{Quora} & 1 & 3 & Get the search results for 'OpenAI'. \\
&  & 2 & 6 & Get the search results for 'OpenAI'. Filter to show only questions. \\
&  & 3 & 11 & Get the search results for 'OpenAI'. Filter to show only questions. Select one question or answer from the results to see more details. Add a comment 'Worth thinking" to the answer. \\
&  & 1 & 3 & Go to Notifications. Check settings for push notifications.  \\
&  & 2 & 8 & Go to Notifications. Check settings for push notifications. Turn off notification from 'Quora Digest' and 'Shares'.   \\
&  & 3 & 11 & Go to Notifications. Check settings for push notifications. Turn off notification from 'Quora Digest'. Return to home. Upvote one of the contents.   \\
&  & 1 & 4 & Get the search results for 'We talk about books'. Return to home. \\
&  & 2 & 6 & Get the search results for 'We talk about books'. Enter 'We talk about books' space. Follow the space. Read one of the posts. \\
&  & 3 & 10 & Get the search results for 'We talk about books'. Follow 'We talk about books' space. Follow the space. Read one of the posts. Follow the post. Return to home. Check 'Following' tab.  \\
\cmidrule{2-5}
& \multirow{3}{*}{BBC} & 1 & 3 & Navigate to 'More'. Get the search results for 'Trump'.  \\
&  & 2 & 9 & Navigate to 'More'. Get the search results for 'Trump'. Enter one of the results. Save this article. Go back to search result. Get the search results for 'Elon Mask'.  \\
&  & 3 & 14 & Navigate to 'More'. Get the search results for 'Trump'. Enter one of the results. Save this article. Go back to search result. Get the search results for 'Elon Mask'. Enter one of the results. Save this article. Go to Saved Items to confirm the articles were added.  \\
\midrule
\multirow{26}{*}{Prod\&Tools}
& \multirow{3}{*}{Chrome} & 1 & 3 & Get the search results for Taylor Swift.  \\
&  & 2 & 10 & Get the search results for Taylor Swift. Go to her Wikipedia page. Add it to bookmarks. Check the Bookmarks for confirmation. \\
&  & 3 & 16 & Get the search results for Taylor Swift. Go to her Wikipedia page. Add it to bookmarks. Move this bookmark to Reading List. Check the Reading List for confirmation. \\
\cmidrule{2-5}

& \multirow{2}{*}{Files} & 2 & 7 & Go to the 'DCIM' folder in the internal storage. Create a subfolder named 'Agent created'.  \\
&  & 3 & 18 & Go to the 'DCIM' folder in the internal storage. Create a subfolder named 'Agent created 2'. Create another subfolder named 'Agent created 3'. Then move the folder 'Agent created 2' into the 'Documents' folder in the internal storage.  \\
\cmidrule{2-5}
& \multirow{3}{*}{Microsoft OneNote} & 1 & 4 & Create a new page with title 'Benchmark' and content 'Test Agent'.  \\
&  & 2 & 7 & Create a new page with title 'Benchmark2' and content TODO 'AppAgent' and 'Mobile Agent'.  \\
&  & 3 & 11 & Create a new notebook 'test'. Create a new section 'prompts' in 'test' notebook. Enter section 'prompts' for confirmation.  \\

\bottomrule
\end{tabular}
}

\end{table*}

\begin{table*}[h!]
\centering
\caption{Single-App Task Instructions (Challenging Task Set) - 5.}
\label{table:challenging_task_set}
\resizebox{\textwidth}{!}{ 
\begin{tabular}{ccccp{0.7\textwidth}}
\toprule
\textbf{Task Category} & \textbf{App Involved} & \textbf{Task Level} & \textbf{Golden Step} & \textbf{Task Instruction} \\
\midrule
\multirow{9}{*}{Shop\&Fin}
& \multirow{6}{*}{Amazon} & 1 & 3 & Get the search results for 'sunglasses'. \\
&  & 2 & 8 & Get the search results for 'sunglasses'. Filter with 'kids'. Add one result to cart. Confirm that this item is in the cart.  \\
&  & 3 & 11 & Get the search results for 'goggles'. Filter with 'adult'. Add one result to cart. Confirm that this item is in the cart. Compare with similar items. Add one of the similar items to cart.  \\
&  & 1 & 4 & Get the search results for 'gaming mouse'. Check details of one result. \\
&  & 2 & 6 & Get the search results for 'gaming mouse'. Check details of two results.  \\
&  & 3 & 10 & Get the search results for 'gaming mouse'. Check details of one result. Go to personal info. Check Amazon prime plan details. \\
\cmidrule{2-5}
& \multirow{3}{*}{Temu} & 1 & 3 & Get the search results for gaming headset. \\
&  & 2 & 7 & Get the search results for gaming headset. Sort the result by the lowest price to highest. Add one to my shopping cart. Confirm that this item is in the cart.  \\
&  & 3 & 12 & Get the search results for gaming mouse. Filter items priced above 10. Add one to cart. Confirm that this item is in the cart. \\

\midrule
\multirow{9}{*}{SystemApps}
& \multirow{3}{*}{Calendar} & 1 & 5 & Check the upcoming 31 October. Click on the event for that day. \\
&  & 2 & 9 & Set up an all-day event titled 'Haircut' on the 16th of any month.  \\
&  & 3 & 14 & Set up an event titled 'Dental Check' on the 17th of any month. Set the time to from 7pm to 9pm.  \\
\cmidrule{2-5}
& \multirow{3}{*}{Contacts} & 1 & 7 & Create a contact named 'Agent'. The phone number is +44 1234 567 890.  \\
&  & 2 & 11 & Create a contact named 'Agent Two'. The phone number is +44 1234 567 890. The email is benchmark@gmail.com  \\
&  & 3 & 15 & Modify the last name of one of the contacts to 'Three'. Update the label for the contact's phone number to Work. Set the company to 'Huawei'. Add an email agent.benchmark.2024@gmail.com. Label the email as Work.  \\
\midrule
\multirow{14}{*}{Travel\&Nav}
& \multirow{2}{*}{Airbnb} & 2 & 8 & Get the search results for stay tonight near 'wembley stadium' for 1 adult.  \\
&  & 3 & 13 & Get the search results for stay tonight near 'wembley stadium' for 1 adult. Add one result to wishlist. Confirm that this item is in the wishlist. \\
\cmidrule{2-5}
& \multirow{3}{*}{Booking} & 1 & 5 & Get the search results for stays in Berlin. Select any date, rooms and guests. \\
&  & 2 & 9 & Navigate to Flights section. Select any date. Choose a flight from Manchester Airport to CDG Paris. Get the search results for a round trip. \\
&  & 3 & 15 & Navigate to Flights section. Select one way flight. Choose the 1st of any month as the flight date. Get the search results from Shanghai to London. \\
\cmidrule{2-5}
& \multirow{6}{*}{Expedia} & 1 & 4 & Check stays in Rome. The dates do not matter. Get the search results for 1 room and 2 people.  \\
&  & 2 & 8 & Check stays in Paris. Choose from 25th to 28th any month. Get the search results for 1 room for 2 people. \\
&  & 3 & 12 & Check stays in Hong Kong. Choose from 25th to 28th any month. Get the search results for 1 room for 2 people. Filter hotels with parking.  \\
&  & 1 & 7 & Check things to do in Paris. Get the search results for 25th to 28th of any month. \\
&  & 2 & 11 & Check things to do in Rome. Get the search results for 26th to 29th of any month. Save it to my trips.  \\
&  & 3 & 13 & Check things to do in Paris. Get the search results for 25th to 28th of any month. Save it to my trips. Confirm that by checking the saved Paris trip.  \\
\bottomrule

\end{tabular}
}

\end{table*}
\begin{table*}[h!]
\centering
\caption{Cross-App Task Instructions.}
\label{tab:cross_app_task_set}
\resizebox{\textwidth}{!}{ 
\begin{tabular}{cccp{0.7\textwidth}}
\toprule
\textbf{Task Category} & \textbf{App Involved} & \textbf{Golden Step} & \textbf{Task Instruction} \\
\midrule
\multirow{3}{*}{\shortstack[l]{General\\Tool}} 
& Clock, Setting & 12 & In the Settings app, enable `Data Saver' mode. Open the Clock app and set an alarm for 6:00 AM.\\
& Keep Notes, LinkedIn & 12 & Use the LinkedIn app to search for a customer service representative position. Select a job, open Keep Notes, create a new note, record the company's name, and set the note's title to `customer service representative'. \\
\midrule
\multirow{3}{*}{\shortstack[l]{Information\\Management}} 
& Facebook, Setting & 17 & Open Facebook, search for tropical pictures, save one picture to your phone, go to the Wallpaper section in the Settings app, and set the saved picture as your wallpaper. \\
& Calendar, Chrome & 16 & Using Chrome, search for the date of the next Winter Olympics opening ceremony and then set a reminder for that date in your Calendar.  \\
& Spotify, Chrome & 13 & Open Chrome, search for the top Country songs of 2023, identify a song from the search results, then switch to Spotify and add that song to your playlist. \\
\midrule
\multirow{3}{*}{\shortstack[l]{Media\\Entertainment}} 
& Clock, Youtube & 11 & Search for a relaxing soundscape video on YouTube, use the Clock app to set a timer for 3 hours, then go back to YouTube and play the video.  \\
\midrule
\multirow{3}{*}{\shortstack[l]{Multi\\Apps}} 
& Quora, eBay, Chrome & 20 & Utilize Chrome to search for a biography book, then use Quora to read reviews about the book, and finally add the book to the watchlist on eBay. \\
& Clock, Chrome, Instagram & 20 & Organize a movie night by choosing a horror film using Chrome, sending an invitation to one of your friends via Instagram, and setting a reminder in the Clock app for 8:35 PM on Sunday. \\
& Clock, WhatsApp, Zoom & 23 & Arrange a business meeting using Zoom, copy the sharing text, go to WhatsApp, send the copied text to a contact, set an alarm using the Clock app at the meeting time. \\
\midrule
\multirow{3}{*}{\shortstack[l]{Social\\Sharing}} 
& X, Facebook & 20 & Use the social media platform X to post a photo, copy the link to your post, then open Facebook and send the link to a friend.  \\
& BBC News, Gmail & 10 & Use the BBC News app to search for Artificial Intelligence news, read an article, share it via Gmail, send to agent.benchmark.2024@gmail.com. \\
& Spotify, Facebook & 19 & Listen to a Reggaeton album on Spotify, then share the album's name with a friend on Facebook.  \\
\midrule
\multirow{3}{*}{\shortstack[l]{Web\\Shopping}} 
& eBay, Facebook & 15 & Search for `Circe by Madeline Miller' on Facebook, read one of the posts, head over to eBay, search for the book, and add it to the watchlist. \\
& Amazon, Temu & 15 & Investigate the prices for Catan board game across Amazon and Temu, then proceed to add the cheaper option into your cart.  \\
& Airbnb, Instagram & 19 & Use Instagram to search for an itinerary for Venice, Italy, and then proceed to Airbnb, book accommodations at Venice, Italy.  \\
\bottomrule
\end{tabular}
}
\end{table*}

\end{document}